\documentclass{article}
\usepackage{spconf,amsmath,epsfig}

\let\OLDthebibliography\thebibliography
\renewcommand\thebibliography[1]{
  \OLDthebibliography{#1}
  \setlength{\parskip}{0pt}
  \setlength{\itemsep}{0pt plus 0.3ex}
}
\usepackage{graphicx}
\usepackage{multirow}
\usepackage{array}
\usepackage{bbding}
\usepackage{threeparttable}
\begin{document}\sloppy

\def\x{{\mathbf x}}
\def\L{{\cal L}}

\title{Hierarchical Graph Convolutional Skeleton Transformer for Action Recognition}
%
\name{Ruwen Bai}
%
%
\address{Institute of Information Engineering, Chinese Academy of Sciences, Beijing, China\\
\textit {bairuwen}@iie.ac.cn}

\maketitle

\begin{abstract}
  Graph convolutional networks (GCNs) have emerged as dominant methods for skeleton-based action recognition.
  However, they still suffer from two problems, namely, neighborhood constraints and entangled spatiotemporal feature representations.
  Most studies have focused on improving the design of graph topology to solve the first problem but they have yet to fully explore the latter.
  In this work, we design a disentangled spatiotemporal transformer (DSTT) block to overcome the above limitations of GCNs in three steps:
  (i) feature disentanglement for spatiotemporal decomposition;
  (ii) global spatiotemporal attention for capturing correlations in the global context;
  and (iii) local information enhancement for utilizing more local information.
  Thereon, we propose a novel architecture, named Hierarchical Graph Convolutional skeleton Transformer (HGCT), to employ the complementary advantages of GCN (i.e., local topology, temporal dynamics and hierarchy) and Transformer (i.e., global context and dynamic attention).
  HGCT is lightweight and computationally efficient.
  Quantitative analysis demonstrates the superiority and good interpretability of HGCT.
\end{abstract}
\section{Introduction}
\label{sec:intro}
Human action recognition has been widely used in many applications, such as video surveillance and smart retail.
With the development of low-cost depth sensors\cite{zhang2012microsoft} and pose estimation algorithms\cite{cao2019openpose}, skeleton-based action recognition has become more attractive.
On this research topic, methods based on graph convolutional networks (GCNs) have emerged as the dominant technology with advanced performance because of their expressive capability of human topology.

Most existing GCN-based methods\cite{yan2018spatial,shi2019two} apply graph convolution to aggregate weighted features of joints according to a predefined graph topology,
followed by temporal convolution to model short-term temporal dynamics.
GCNs thus encode hierarchical spatiotemporal features to characterize actions by stacking multiple spatiotemporal graph convolution operations (STGC).
However, this operation mechanism limits the upper bound of GCNs.
First, feature aggregation is restricted in the local spatial–temporal neighborhood.
Second, stacked STGC operations jointly learn spatiotemporal features.
As such, the spatiotemporal feature learning of a network is essentially a black box. 
To solve the first problem, researchers have improved the design of the graph topology to strengthen the power of STGC.
But another problem is often ignored in the task of skeleton-based action recognition.
Unlike previous work, we employ Transformer to relax the local constraints from GCNs based on its strong capability of global modeling.
The other problem is studied due to the consideration of computation efficiency.

Transformer proposed by\cite{vaswani2017attention} achieves excellent performance because of its dynamic attention to all input tokens.
But regardless of which method all has its dual character.
Transformer is insensitive to the local context, and its good performance always depends on increasing iteration times or sufficient data.
This is because Transformer assumes minimal inductive biases.
Then, look back to GCNs, they have obvious advantages in modeling local topology, hierarchical semantics and short-term temporal information.
Therefore, with these characteristics, we determine that Transformer and GCNs have complementary strengths and enjoy great potentials for cooperation.

\begin{figure*}[t]
  \centering
  \includegraphics[width=0.98\textwidth]{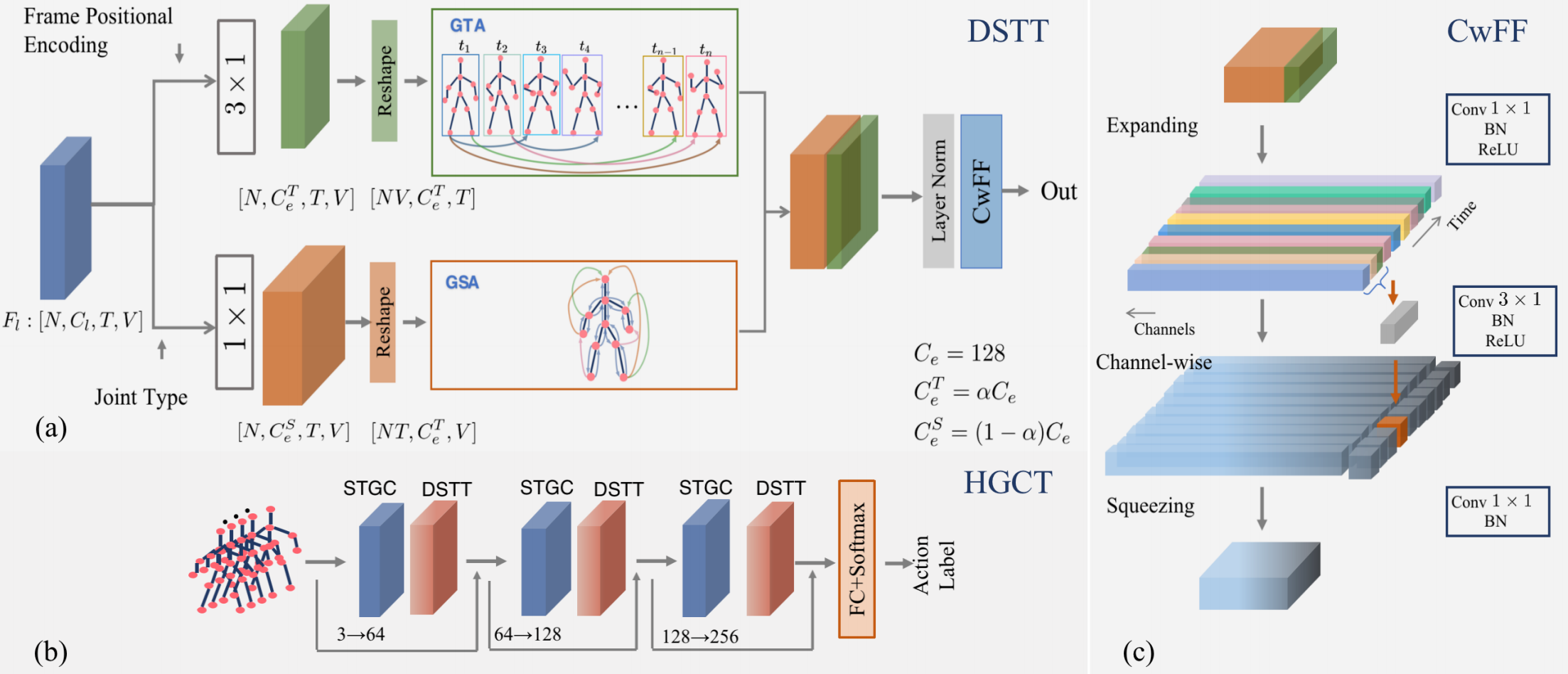} 
  \caption{\textbf{Architecture Overview.} (a) Illustration of the disentangled spatiotemporal transformer (DSTT) block. (b) Illustration of the hierarchy design of HGCT. (c) Illustration of the channel-wise feed forward module (CwFF).}
  \label{pipeline}
\end{figure*}
To maximize their strengths, we design a disentangled spatiotemporal transformer (DSTT) block, as shown in Figure~\ref{pipeline}(a).
The DSTT block operates in three steps:
(i) feature disentanglement to decompose spatial and temporal features;
(ii) global spatiotemporal attention to pay dynamic attention to all joints and frames of the skeleton sequence;
and (iii) local information enhancement to remain more local information.
On this basis, a novel architecture called Hierarchical Graph Convolutional Skeleton Transformer (HGCT) is presented, as shown Figure~\ref{pipeline}(b).
By applying STGC, we first introduce multi-level action patterns as appropriate inductive biases, which help accelerate the convergence.
Then via feature disentanglement, self-attention operations in the DSTT specialize in separated spatial and temporal component to model the global context.
Thus, computing becomes more efficient.
Through the channel-wise feed forward module (CwFF), local information can be disseminated effectively across layers. 
Furthermore, our model improves the interpretability of spatial–temporal feature representation.
The main contributions of this work are as follows:
\begin{itemize}
  \item A variant of Transformer block, DSTT is designed to provide strong complementarities to GCNs. Thus, effective representations in the local–global context can be achieved.
  \item A novel architecture called HGCT is presented to exploit all the benefits of Transformer (i.e. dynamical attention and global context) while keeping the advantages of GCNs (i.e. hierarchy, local topology and temporal dynamics).
  \item With a well-designed architecture, HGCT offers all desirable properties, and it is explanatory, lightweight and computationally efficient.
  \item The contributions of spatial and temporal features in different layers are quantitatively analyzed. Exhaustive ablation experiments are conducted to verify the effectiveness of HGCT.
\end{itemize}
\section{Proposed Method}
\label{sec:methods}
As shown in Figure~\ref{pipeline}(b), the overall architecture of HGCT is divided into three stages, which consist of STGC and DSTT.
Residual connections ensure the transmission of local topological information.
In the first stage, the joint type and frame orders are introduced into DSTT through the positional encoding of Transformer.
In this section, we introduce each component of our HGCT.
\subsection{Spatiotemporal GCN}
As a spatiotemporal feature extractor,
STGC stacks a spatial module to model the graph topology and a temporal module to model temporal dynamics.
Hierarchical STGC extracts feature representations with various levels, from local spatial–temporal contexts to high-level action patterns.
\paragraph*{Modeling of spatial topology.}
Raw skeleton sequences are denoted as $X_{in} \in R^{C_{in} \times T\times N}$,
where each vector $X_{in}^{t}=\{v_{1}^{t},v_{2}^{t},\dots, v_{n}^{t}\}$ are the 2D or 3D coordinates of $N$ human joints at a time stamp $t$; 
and $C_{in}$ is 3. The graph convolution operation for each frame is formulated as:
$$f^{l+1}=\sum_{k}^{k_{v}}W^{l}f^{l}( \lambda^{l}\tilde{A}_{k}^{l})$$.
where $f^{l}$ and $f^{l+1}$ denote the input and output feature maps of the $l$th layer, respectively.
In accordance with the partition strategy implemented in a previous study\cite{yan2018spatial}, $K_{v}$ is set to $3$.
The learnable weight $W^{l}$ is unique for each layer. The adjacency matrix $\tilde{A}_{k}$ represents the graph topology, which is initialized with a human-body-based graph.
Unlike the adjacency matrix defined in ST-GCN\cite{yan2018spatial}, $\tilde{A}_{k}$ is parameterized and updated in the training process.
A balanced factor $\lambda$ is introduced with the initial value 1, which learns individually at each layer.
The contribution of local topology to the final output is fine-tuned hierarchically. Thus, STGC relaxes the constraints of graph convolution and allows flexible modeling of graph topology.
\paragraph*{Modeling of temporal dynamics}
STGC models temporal dynamics with multi-scale learning following \cite{liu2020disentangling}.
A multiscale temporal convolution module is composed of two $5\times 1$ temporal convolutions with different dilation rates, 
one $3\times 1$ temporal pooling, 
and one $1\times 1$ bottleneck.
In the first three branches use $1\times 1$ convolution is utilized to reduce the channel dimension.
With this multibranch low-computational-cost design, rich temporal dynamics can be modeled.
\subsection{Disentanglement Spatiotemporal Transformer}
DSTT is a standard Transformer variation.
In this study, we perform three modifications to adapt Transformer for the skeleton-based action recognition.
\begin{figure}[t]
  \centering
  \includegraphics[width=\columnwidth]{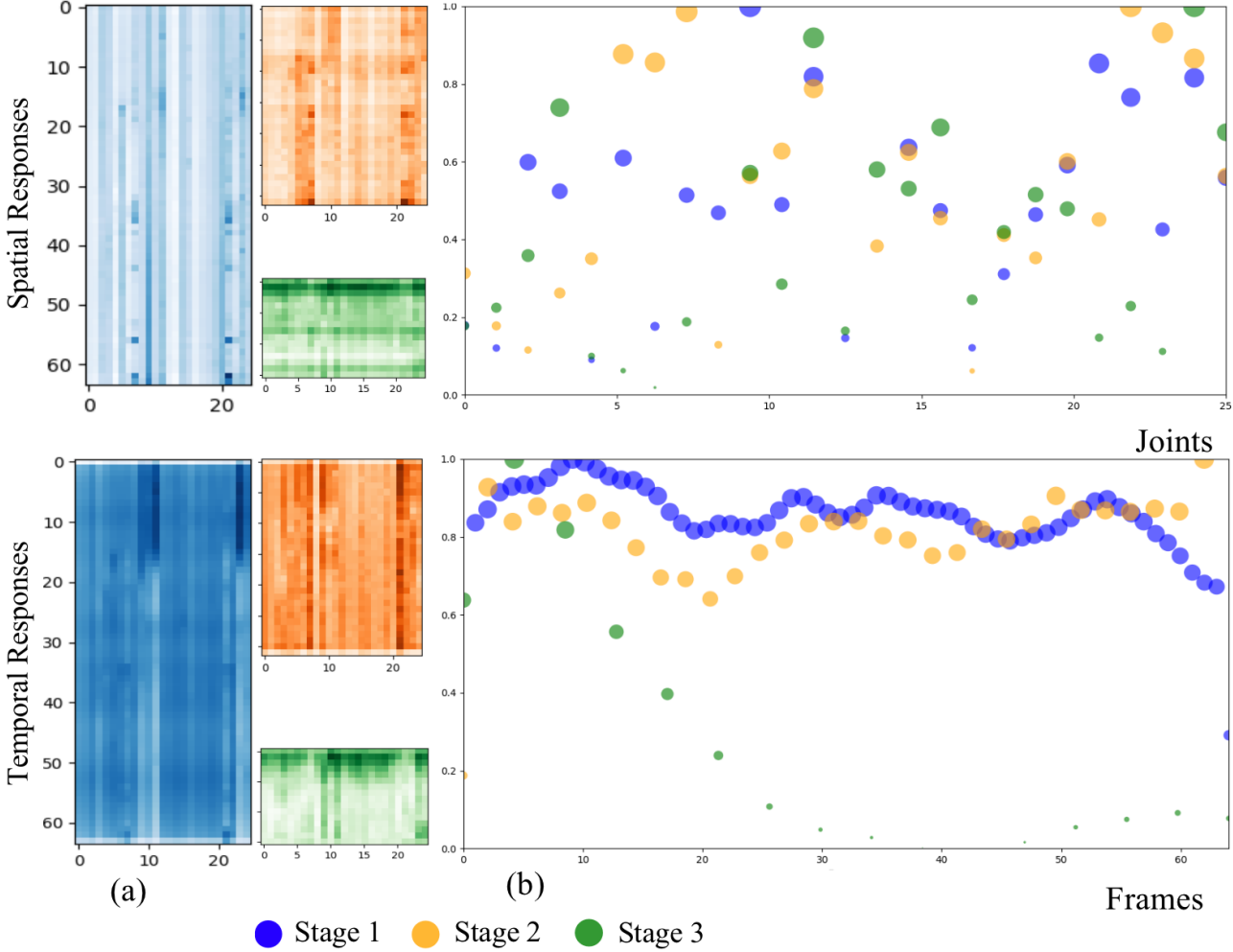} 
  \caption{Visualization of disentangled features at different stages (marked with colors). Top: Spatial responses of joints; Bottom: Temporal responses of frames. (a): the X and Y coordinates represent the indexes of joints and frames, respectively. (b): The Y-coordinate denotes the intensity of feature responses after normalization (shown as the point size).}
  \label{stml}
  \vspace{-0.1cm}
\end{figure}
\paragraph*{Feature disentanglement}
Human poses and joint movements are important cues for human action recognition.
Most studies\cite{yan2018spatial,liu2020disentangling} have explored entangled spatiotemporal features.
In another study\cite{10.1007/978-3-030-89370-5_12}, the skeleton sequences of different frame rates are fed into a multistream architecture.
Each stream concentrates on different spatial and temporal properties.
In our work, spatial and temporal representations are considered separately in one stream by feature disentanglement along the channels.
Spatial features tend to be more discriminative, so two groups of features with different numbers of channels are generated through feature disentanglement.
Given the feature maps $F_{l}\in R^{N \times C_{l} \times T \times V}$, a $3\times 1$ convolution is applied to produce a group of lower-dimensional feature maps $ F_{l}^{T}\in R^{N \times C_{e}^{T} \times T \times V}$,
which are sensitive to joint dynamics in time.
Another group of higher-dimensional feature maps $F_{l}^{S}\in R^{N \times C_{e}^{S} \times T \times V}$ generated by a $1\times 1$ convolution characterize the detailed joint features in space.
This treatment can be explained by the intrinsic mechanism of convolution.
A large number of filters would be used to embed features in high-dimensional representation space.
Thus, detailed spatial information can be extracted.
By contrast, the ability of structure representations is weakened in the low-dimensional feature space.
This point is further supported by the the visualization analysis in Figure~\ref{stml}(a).
Furthermore, lower-dimensional feature maps present apparent changes along temporal channels.
Figure~\ref{stml}(b) shows that low-level features focus on more temporal dynamics.
All stages produce high responses to the spatial semantics, which differ at each other.
Here, we introduce $\alpha$ to specify the contributions of spatial and temporal features.
We also conduct experiments with $\alpha=1/2, 1/4, 1/8$ respectively.
Here, the embedding dimensions $C_{e}=C_{e}^{S}+C_{e}^{T}=128$.

\paragraph*{Global spatiotemporal self-attention}
The spatiotemporal features are decomposed into spatial features $F_{l}^{S}$ and temporal features $F_{l}^{T}$ with the method described above.
In this process, a global spatial attention module (GSA) and a global temporal attention module (GTA) perform multi-head self-attention (MHSA) simultaneously to learn global spatial–temporal correlations.
The former operates in space, the latter in time. MHSA receives a sequential data as input, i.e., $X\in R^{N\times Tokens \times C}$.
$N$, $Tokens$, $C$ denotes the batch size, the number of input tokens and the number of channels, respectively.
To satisfy the input of MHSA, the disentangled spatial features $F_{l}^{S}$ are first reshaped into joint tokens $F_{l}^{J}\in R^{NT \times V\times C_{e}^{S}}$,
while temporal features $F_{l}^{T}$ are reshaped into frame tokens $F_{l}^{F}\in R^{NV \times T\times C_{e}^{T}}$.
GSA and GTA receives joint tokens and frame tokens as input, respectively. 
The computation process in GSA and GTA is the same as MHSA in standard Transformer.
More details are given in the supplemental material.

\paragraph*{Local information enhancement}
The standard transformer disregards the local context even though its attention mechanism is effective in modeling long-range dependencies.
locality is extracted by extending a convolutional FFN version. \cite{li2021localvit} extracts locality by extending a convolutional version FFN.
On this basis, we design a channel-wise feed forward module (CwFF) to enhance the local connectivity.
The output feature map of CwFF is generated through expansion, channel-wise excitation and squeezing.
As shown in Figure~\ref{pipeline}(c), the layer-normalizatized feature map $F_{norm} \in R^{N\times TV \times  C_{e}}$ are first reshaped into $F\in R^{N\times C_{e}\times T\times V}$, then fed into CwFF.
In CwFF, a $1\times 1$ point-wise convolution is first used to obtain higher-dimensional representations,
where the expanding factor $\gamma$ is $3$.
Then the channel-wise convolution with a kernel size $3\times 1$ is conducted on each input channel of feature map.
This process would bring benefits.
First, each output channel of the feature map responds to specific spatial or temporal aspect.
Intuitively, different channels have their corresponding emphases.
For instances, some channels focus on more static pattern in space, and the others concentrate on more motion dynamics.
Second, local temporal information is further aggregated to propagate forward.
Third, the channel-wise convolution is efficient both in terms of parameters and computation.
In the end, another $1\times 1$ point-wise convolution is applied to squeeze the excitated feature map.
point-wise convolution is applied to squeeze the excitated feature map.

\paragraph*{Joint Type and Frame Order}
Through the positional encoding of Transformer, joint type and frame order can be introduced without any efforts.
This process is performed only at the first stage because hierarchical architecture and residual connections are sufficient to convey the information flow. 
$C_{e}^{S}$ dimensions of feature representations for each joint type are embedded, thus $J_{embed} \in R^{N\times C_{e}^{S}}$.
The frame position in skeleton sequences represents the frame order in time, which involves  high-level semantics related to actions.
Sinusoidal positional encoding\cite{vaswani2017attention} is utilized to obtain the order context of frames effortlessly.
The encoding dimensions are $C_{e}^{T}$.
\begin{figure}[t]
  \centering
  \includegraphics[width=\columnwidth]{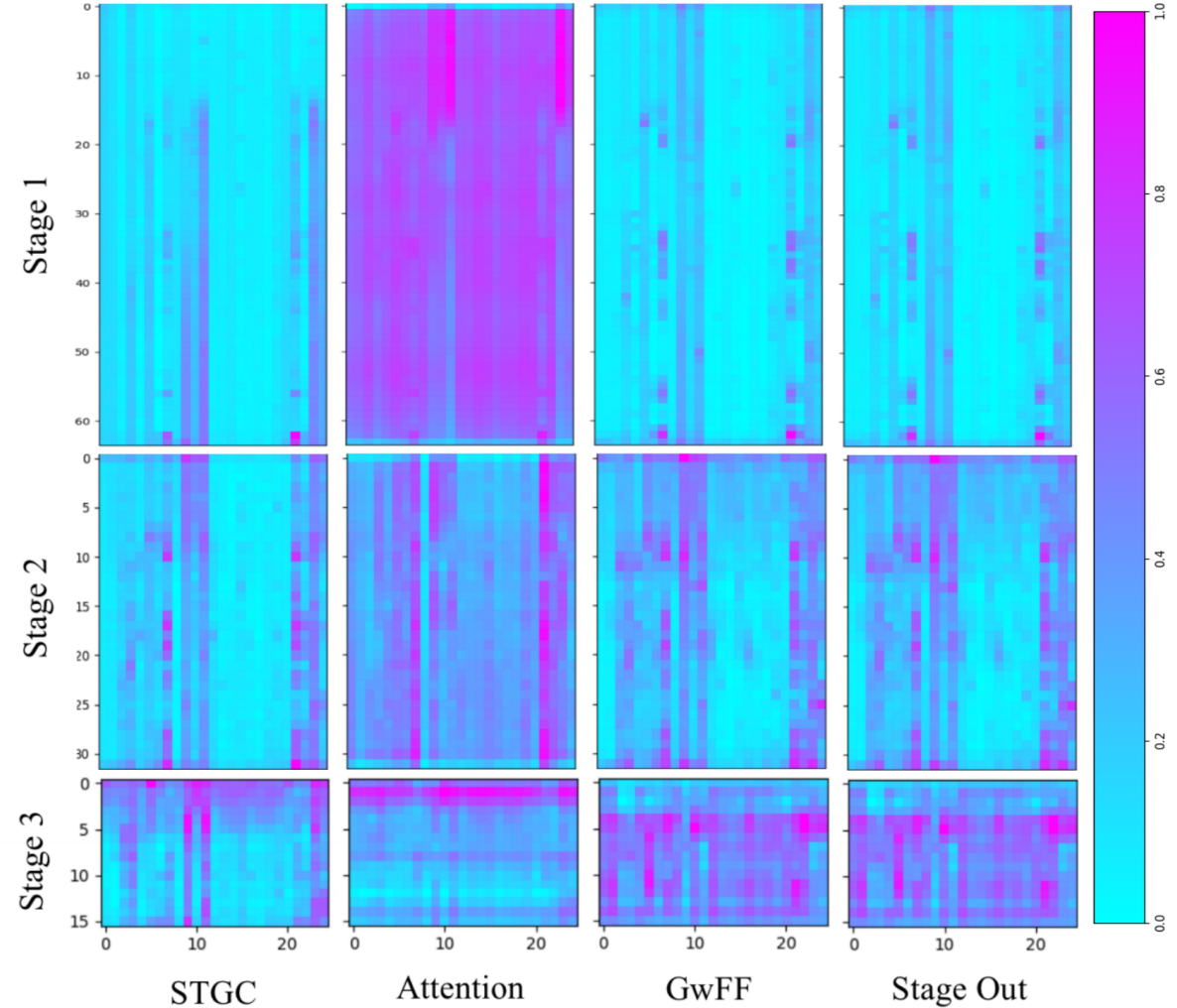} 
  \caption{Visualization of the output features from different blocks. Different rows shows various stages in HGCT.}
  \label{ffn}
\end{figure}
\subsection{Discussion with ST-TR}
ST-TR\cite{plizzari2021spatial} initially employs self-attention operator of Transformer to overcome the restrictions of local neighborhoods in GCNs.
A two-stream framework is adopted in ST-TR, one for combining spatial graph convolution and the other for combining spatial self-attention operations with temporal convolution.
Another key part of Transformer, the feed-forward network (FFN) is removed in ST-TR for saving computation cost and parameters.
Our HGCT differs from it in three respects: First, HGCT introduces global spatiotemporal context in one stream. 
Second, all benefits from Transformer are inherited efficiently by HGCT.
The feature visualization in Figure~\ref{ffn} reveals that CwFF as a variant of FFN detects action-related semantic patterns in a bottom-up manner.
Third, HGCT is lightweight with low-cost computation.

\section{Experiments}
\label{sec:experiments}
\subsection{Datasets}
\par\textbf{NTU RGB+D (NTU 60)} \cite{shahroudy2016ntu} is a human action dataset with 60 categories.
It collects 56,880 video clips from 40 subjects and provides the skeleton data containing 3D joint coordinates of 25 body joints.
Two benchmarks for evaluation are Cross-subject (X-Sub) and Cross-view (X-View).

\par\noindent\textbf{NTU RGB+D 120 (NTU 120)} \cite{liu2019ntu} extends NTU 60 to 120 classes with an additional 57,367 skeleton sequences.
Totaling 113,945 samples are captured by 32 different camera setups and 106 volunteers.
It currently becomes the largest skeleton dataset. Two benchmarks for evaluation are Cross-subject (X-Sub) and Cross-setup (X-Set).
\begin{table}[t!]
  \centering
  \setlength{\tabcolsep}{0.7mm}
  \footnotesize
  \caption{Detailed settings of our HGCT.}
  \vspace{0.1cm}
  \label{settings}
  \begin{tabular}{c|cc|ccc|ccc|cc}
  \hline
  \multirow{2}{*}{Model} & \multicolumn{2}{c|}{DSTT}           & \multicolumn{3}{c|}{GSA}      & \multicolumn{3}{c|}{GTA}      & \multicolumn{2}{c}{CwFF}  \\ \cline{2-11}
                         & \multicolumn{1}{c|}{$C_{e}$} & $\alpha$   & \multicolumn{1}{c|}{$C_{e}^{S}$} & \multicolumn{1}{c|}{$S_{heads}$} &attn drop& \multicolumn{1}{c|}{$C_{e}^{T}$} & \multicolumn{1}{c|}{$T_{heads}$} &attn drop& \multicolumn{1}{c|}{$\gamma$}  &drop                                      \\ \hline
  HGCT                 & \multicolumn{1}{c|}{128}      & 1/4 & \multicolumn{1}{c|}{96}  & \multicolumn{1}{c|}{6}  &0.0& \multicolumn{1}{c|}{32}  & \multicolumn{1}{c|}{8}  & 0.0 & \multicolumn{1}{c|}{3}      &0.0                \\ \hline
  \end{tabular}
  \vspace{-0.1cm}
\end{table}

\subsection{Implementation Details}
\par\noindent\textbf{Network Architecture.}
The detail settings of HGCT are presented in Table~\ref{settings}.
In our experiments, the default settings in Table~\ref{settings} are applied unless otherwise indicated.
$S_{heads}$ and $T_{heads}$ denote the number of attention heads of GSA and GTA, respectively.
We set the dropout ratios to 0 for each module of DSTT.
\par\noindent\textbf{Training.}
All experiments are conducted on two RTX TITIAN GPUs with the PyTorch framework.
We use SGD with momentum (0.9) to train the model for 60 epochs.
Cross entropy loss is utilized with 0.1 label smoothing rate.
The initial learning rate of 0.05 is reduced by 10 at the 40th and 50th epochs.
A linear warmup training\cite{shi2019two} is implemented in the first 5 epochs.
The weight decay and the batch size are set to 0.0002 and 64, respectively.
Raw skeleton sequences are downsampled to a fixed size of 64 frames.
Data preprocessing is conducted with the same strategy as introduced in\cite{li2019spatio}.

\begin{table*}[t!]
  \centering
  \footnotesize
  \caption{Comparison with state-of-the-art accuracy (\%) on NTU 60/120.}
    \label{sota}
    \vspace{0.1cm}
    \begin{threeparttable}
    \setlength{\tabcolsep}{3mm}
    \small
    \begin{tabular}{c|c|c|cc|cc|c|c}
    \hline
    \multirow{2}{*}{Method} & \multirow{2}{*}{\begin{tabular}[c]{@{}c@{}}Data\\ Modalities\end{tabular}}                                                                           & \multirow{2}{*}{Model}       & \multicolumn{2}{c|}{NTU 60}                    & \multicolumn{2}{c|}{NTU 120}                       & \multirow{2}{*}{Params(M)}         &  \multirow{2}{*}{FLOPs(G)}                 \\ \cline{4-7}
                            &                                                                            &               & \multicolumn{1}{c|}{X-Sub}         & X-View    & \multicolumn{1}{c|}{X-Sub}         & X-Set         &  &           \\ \hline
    \multirow{6}{*}{GCNS}   &\multirow{2}{*}{Joint}                                                      & ST-GCN\cite{yan2018spatial}        & \multicolumn{1}{c|}{81.5}          & 88.3      & \multicolumn{1}{c|}{$70.7^{\star}$}          & $73.2^{\star}$          & $3.10^{\star}$       & $\sim3.56^{\star}$         \\
    &                                                                            & AS-GCN\cite{li2019actional}       & \multicolumn{1}{c|}{86.8}          & 94.2      & \multicolumn{1}{c|}{$77.9^{\star}$}          & $78.5^{\star}$          & $6.94^{\star}$      & $\sim6.10^{\star}$         \\ \cline{2-9}
    &\multirow{2}{*}{\begin{tabular}[c]{@{}c@{}}Joint+\\ Bone\end{tabular}}      & 2s-AGCN\cite{shi2019two}       & \multicolumn{1}{c|}{88.5}          & 95.1      & \multicolumn{1}{c|}{$82.5^{\star}$}          & $84.2^{\star}$          & $6.94^{\star}$      & $\sim7.96^{\dagger}$                  \\
    &                                                                            & MS-G3D\cite{liu2020disentangling}        & \multicolumn{1}{c|}{91.5}          & 96.2      & \multicolumn{1}{c|}{86.9}          & 88.4          & 2.8       & $\sim8.32^{\star}$         \\ \cline{2-9}
    & \multirow{2}{*}{\begin{tabular}[c]{@{}c@{}}Four \\ modalities\end{tabular}}                                                                           & 4s-Shift GCN\cite{cheng2020skeleton}  & \multicolumn{1}{c|}{90.7}          & 96.5      & \multicolumn{1}{c|}{85.9}          & 87.6          & —         & \textbf{$\sim$0.7} \\
    &                                                                            & Dynamic GCN\cite{ye2020dynamic}   & \multicolumn{1}{c|}{91.5}          & 96.0      & \multicolumn{1}{c|}{87.3}          & 88.6          & —         & $\sim$1.99         \\ \hline
    \multirow{7}{*}{Transformers} &\multirow{3}{*}{Joint}                                                      & ST-TR\cite{plizzari2021spatial}         & \multicolumn{1}{c|}{88.7}          & 95.6      & \multicolumn{1}{c|}{81.9}          & 84.1          & $23.5^{\dagger}$      & $\sim27.8^{\dagger}$         \\
    &                                                                            & ST-TR-agcn    & \multicolumn{1}{c|}{89.2}          & 95.8      & \multicolumn{1}{c|}{82.7}          & 85.0          & —         & —                  \\
    &                                                                            & HGCT(Ours)          & \multicolumn{1}{c|}{89.7}          & 95.3      & \multicolumn{1}{c|}{85.4}          & 87.1          & 0.93      & $\sim$1.5          \\ \cline{2-9}
    &\multirow{3}{*}{\begin{tabular}[c]{@{}c@{}}Joint+\\ Bone\end{tabular}}      & ST-TR         & \multicolumn{1}{c|}{89.9}          & 96.1      & \multicolumn{1}{c|}{84.3}          & 86.7          & —         & —                  \\
    &                                                                            & ST-TR-agcn    & \multicolumn{1}{c|}{90.3}          & 96.3      & \multicolumn{1}{c|}{85.1}          & 87.1          & —         & —                  \\
    &                                                                            & HGCT(Ours)          & \multicolumn{1}{c|}{91.6} & 96.2      & \multicolumn{1}{c|}{88.9}          & 90.0          & 1.85      & $\sim$3.0          \\ \cline{2-9}
    &\begin{tabular}[c]{@{}c@{}}Four\\ modalities\end{tabular}                   & HGCT(Ours) & \multicolumn{1}{c|}{\textbf{92.2}}     & \textbf{96.5} & \multicolumn{1}{c|}{\textbf{89.2}} & \textbf{90.6} & 3.70      & $\sim$4.5          \\ \hline
    \end{tabular}
    \begin{tablenotes}    
      \footnotesize               
      \item $\star$: These results are provide in \cite{song2020stronger,ye2020dynamic}; $\dagger$: These results are calculated according to their released codes.
      \item Input skeleton sequence is $64$ frames $\times$ $25$ joints by default for computing FLOPs ($\times 10^{9}$).
    \end{tablenotes}            
    \end{threeparttable}
    \end{table*}
\subsection{Comparison with the State-of-the-arts}
To validate the effectiveness of HGCT, we first compare HGCT models with GCN-based methods, which have dominated the skeleton-based action recognition.
Then we select the representative work ST-TR for comparison to further demonstrate the design of HGCT.
Many state-of-the-art methods are extended to multi-stream frameworks for different data modalities, including joint, bone, joint motion and bone motion.
In this study, we report the performance in the same way for a fair comparison (shown in Table~\ref{sota}).
On both NTU 60 and NTU 120 datasets, our HGCT achieves superior results under all evaluation.
\textit{v.s. GCNs:} HGCT shows a competitive performance when only joint information is used.
When trained on the same data modality, HGCT exceeds other methods by a large margin with far fewer parameters.
\textit{v.s. ST-TR:} When joint information and bone information are used, HGCT outperforms ST-TR by 4.6\% and 3.3\%, respectively, on X-Sub and X-Set benchmarks of NTU 120. 
Notably, our HGCT is a great lightweight architecture that drastically reduces the computational cost of ST-TR.
On the basis of these results, we can draw \textit{three main conclusions}:
First, the generalization capability of HGCT for large datasets is better than that of GCN-based methods.
Second, HGCT doesn not rely on the intricate designs of topology structure by playing through the complementary strengthens between Transformer and GCNs.
Third, as a Transformer-based model, HGCT achieves a good trade-off between accuracy and computational complexity.
\subsection{Ablation Study}
We conduct ablation experiments to verify the design of HGCT. Unless stated otherwise, performance is reported as classification accuracy(\%) on the X-Set benchmark of NTU120 by using the joint data only.

\begin{table}[t]
  \vspace{-0.3cm}
  \caption{Ablation Study on NTU 120 X-Set.}
  \label{ablation}
  \begin{minipage}[b]{0.45\linewidth}
      \centering
      \vspace{0.2cm}
      \footnotesize
      \centerline{(a) Topology learning}
      \setlength{\tabcolsep}{1.2mm}
      \begin{tabular}{c|c|c|c}
      \hline
      Topology & $A_{o}$   & $\tilde{A}$   & $\lambda\tilde{A}$            \\ \hline
      Accuracy & 85.7 & 86.8 & \textbf{87.1} \\ \hline
      \end{tabular}
    \end{minipage}
    \begin{minipage}[b]{0.45\linewidth}
      \centering
      \vspace{0.2cm}
      \footnotesize
      \centerline{(b) Proportions of feature disentanglement}
      \setlength{\tabcolsep}{1.5mm}
      \footnotesize
      \begin{tabular}{c|c|c|c}
        \hline
        alpha    & 1/2  & 1/4           & 1/8           \\ \hline
        Accuracy & 86.7 & \textbf{87.1} & 86.6 \\ \hline
      \end{tabular}
    \end{minipage}
    \vfill
    \begin{minipage}[b]{0.9\linewidth}
      \centering
      \vspace{0.2cm}
      \footnotesize
      \centerline{(c) Effect of the expanding factor}
      \setlength{\tabcolsep}{4.5mm}
      \footnotesize
      \begin{tabular}{c|c|c|c|c}
        \hline
        $\gamma$         & 1.0  & 2.0  & 3.0           & \multicolumn{1}{c}{4.0} \\ \hline
        Params(M) & 0.72 & 0.82 & 0.92          & \multicolumn{1}{c}{1.0} \\ \hline
        Accuracy  & 86.1     & 86.5 & \textbf{87.1} & 86.5                     \\ \hline
      \end{tabular}
    \end{minipage}
    \vfill
    \begin{minipage}[b]{0.9\linewidth}
      \centering
      \vspace{0.2cm}
      \footnotesize
      \centerline{(d) Effect of joint type and frame order}
      \setlength{\tabcolsep}{2.1mm}
      \footnotesize
      \begin{tabular}{c|c|c|c|c|c} 
        \hline
        Method & \begin{tabular}[c]{@{}c@{}}Joint \\ Type\end{tabular} & \begin{tabular}[c]{@{}c@{}}Frame \\ Order\end{tabular} & Param(M) & FLOPs     & Acc(\%) \\\hline
        a      &                                                       &                                                        & 9.2         &  $\sim$1.5         & 86.8    \\ \hline
        b      &                                                       & \Checkmark                                                     & 9.2      & $\sim$1.5 & 86.3   \\ \hline
        c      & \Checkmark                                                     &                                                        & 9.2      & $\sim$1.5 & 86.7   \\ \hline
        d      & \Checkmark                                                     & \Checkmark                                                      & 9.2      & $\sim$1.5 & 87.1 \\ \hline
        \end{tabular}
    \end{minipage}
  \end{table}
\noindent\textbf{Local Topology Learning}
The flexibility of topology learning in STGC determines whether the local topology information is effectively introduced into the Transformer block.
Therefore, we investigate different settings of the adjacency matrix.
Here we denote the original adjacency matrix based on the intrabody connections as $A_{o}$.
(1) We fix the original adjacency matrix across layers. (marked as $A_{o}$ in );
(2) Similar to (1), we parameterize and initialize the adjacency matrix $\tilde{A}$ with $A_{o}$, which is fixed at the first 5 epoch(marked as $\tilde{A}$);
(3) We multiply the balance factor $\lambda$ with the initial value of 1 by the adjacency matrix $\tilde{A}$ (marked as $\lambda\tilde{A}$).
The results are shown in Table~\ref{ablation}(a).
The balance factor further adjusts the contributions of local information at each layer. 

\par\noindent\textbf{Feature Disentanglement}
We investigate the effects of variations in the proportion of spatial and temporal channels after disentanglement.
The results are shown in Table~\ref{ablation}(b).
$\alpha$ means the temporal channel ratio of embedding channels.
Our results show that a lower channel capacity does not hinder the modeling of temporal dynamics.
The feature visualizations shown in Figure~\ref{stml} and Figure~\ref{ffn} help further elucidate the interaction of spatial and temporal features. 

\par\noindent\textbf{Expanding in CwFF}
We explore the different configurations of the expanding factor in Table~\ref{ablation}(c).
Interestingly, the accuracy is not improved even though $\gamma$ increases.
When the expanding factor is set to 3, HGCT achieves the best performance HGCT achieves the best performance possibly because excessively high hidden dimensions cause overfitting.
\par\noindent\textbf{Joint Type and Frame order}
As shown in Table~\ref{ablation}(d), we introduce joint type and frame order without extra parameters and computation and still improve the performance.
The absence of joint type likely leads to a significant drop in accuracy.
As mentioned in Section~\ref{sec:methods}, joint features have more discriminative information beneficial to action recognition.
\section{Conclusion}
In this work, we present a novel architecture HGCT for skeleton-based action recognition.
We design a Transformer block variant, i.e., DSTT, which is incorporated into GCNs in an efficient and plug-and-play manner.
DSTT takes into account both global context and local information, thus providing strong complementarities with GCNs.
Our work first attempts to disentangle spatial and temporal features and consequently making possible interpretations for skeleton-based action recognition.
Besides, without placing an extra burden on the model, we introduce the joint type and frame orders to help HGCT comprehensively understand human actions.
What's more, HGCT does not have a complicated topology design, but still achieves advanced performance with few parameters and low computational cost.
Our detailed discussion and visualization demonstrated that HGCT gives full play to the advantages of Transformer and GCNs.
Overall, this study promotes the development of Transformer-based models on this task.

\bibliographystyle{IEEEbib}
\bibliography{icme2022template}

\begin{thebibliography}{10}

\bibitem{zhang2012microsoft}
Zhengyou Zhang,
\newblock ``Microsoft kinect sensor and its effect,''
\newblock {\em IEEE multimedia}, , no. 2, pp. 4--10, 2012.

\bibitem{cao2019openpose}
Zhe Cao, Gines Hidalgo, Tomas Simon, Shih-En Wei, and Yaser Sheikh,
\newblock ``Openpose: realtime multi-person 2d pose estimation using part
  affinity fields,''
\newblock {\em IEEE transactions on pattern analysis and machine intelligence},
  vol. 43, no. 1, pp. 172--186, 2019.

\bibitem{yan2018spatial}
Sijie Yan, Yuanjun Xiong, and Dahua Lin,
\newblock ``Spatial temporal graph convolutional networks for skeleton-based
  action recognition,''
\newblock in {\em Proceedings of the AAAI conference on artificial
  intelligence}, 2018.

\bibitem{shi2019two}
Lei Shi, Yifan Zhang, Jian Cheng, and Hanqing Lu,
\newblock ``Two-stream adaptive graph convolutional networks for skeleton-based
  action recognition,''
\newblock in {\em Proceedings of the IEEE/CVF Conference on Computer Vision and
  Pattern Recognition}, 2019, pp. 12026--12035.

\bibitem{vaswani2017attention}
Ashish Vaswani, Noam Shazeer, Niki Parmar, Jakob Uszkoreit, Llion Jones,
  Aidan~N Gomez, {\L}ukasz Kaiser, and Illia Polosukhin,
\newblock ``Attention is all you need,''
\newblock in {\em Advances in neural information processing systems}, 2017, pp.
  5998--6008.

\bibitem{liu2020disentangling}
Ziyu Liu, Hongwen Zhang, Zhenghao Chen, Zhiyong Wang, and Wanli Ouyang,
\newblock ``Disentangling and unifying graph convolutions for skeleton-based
  action recognition,''
\newblock in {\em Proceedings of the IEEE/CVF Conference on Computer Vision and
  Pattern Recognition}, 2020, pp. 143--152.

\bibitem{10.1007/978-3-030-89370-5_12}
Ruwen Bai, Xiang Meng, Bo~Meng, Miao Jiang, Junxing Ren, Yang Yang, Min Li, and
  Degang Sun,
\newblock ``Graph attention convolutional network with motion tempo enhancement
  for skeleton-based action recognition,''
\newblock in {\em PRICAI 2021: Trends in Artificial Intelligence}, Duc~Nghia
  Pham, Thanaruk Theeramunkong, Guido Governatori, and Fenrong Liu, Eds., 2021,
  pp. 152--165.

\bibitem{li2021localvit}
Yawei Li, Kai Zhang, Jiezhang Cao, Radu Timofte, and Luc Van~Gool,
\newblock ``Localvit: Bringing locality to vision transformers,''
\newblock {\em arXiv preprint arXiv:2104.05707}, 2021.

\bibitem{plizzari2021spatial}
Chiara Plizzari, Marco Cannici, and Matteo Matteucci,
\newblock ``Spatial temporal transformer network for skeleton-based action
  recognition,''
\newblock in {\em International Conference on Pattern Recognition}. Springer,
  2021, pp. 694--701.

\bibitem{shahroudy2016ntu}
Amir Shahroudy, Jun Liu, Tian-Tsong Ng, and Gang Wang,
\newblock ``Ntu rgb+ d: A large scale dataset for 3d human activity analysis,''
\newblock in {\em Proceedings of the IEEE conference on computer vision and
  pattern recognition}, 2016, pp. 1010--1019.

\bibitem{liu2019ntu}
Jun Liu, Amir Shahroudy, Mauricio Perez, Gang Wang, Ling-Yu Duan, and Alex~C
  Kot,
\newblock ``Ntu rgb+ d 120: A large-scale benchmark for 3d human activity
  understanding,''
\newblock {\em IEEE transactions on pattern analysis and machine intelligence},
  vol. 42, no. 10, pp. 2684--2701, 2019.

\bibitem{li2019spatio}
Bin Li, Xi~Li, Zhongfei Zhang, and Fei Wu,
\newblock ``Spatio-temporal graph routing for skeleton-based action
  recognition,''
\newblock in {\em Proceedings of the AAAI Conference on Artificial
  Intelligence}, 2019, vol.~33, pp. 8561--8568.

\bibitem{li2019actional}
Maosen Li, Siheng Chen, Xu~Chen, Ya~Zhang, Yanfeng Wang, and Qi~Tian,
\newblock ``Actional-structural graph convolutional networks for skeleton-based
  action recognition,''
\newblock in {\em Proceedings of the IEEE/CVF Conference on Computer Vision and
  Pattern Recognition}, 2019, pp. 3595--3603.

\bibitem{cheng2020skeleton}
Ke~Cheng, Yifan Zhang, Xiangyu He, Weihan Chen, Jian Cheng, and Hanqing Lu,
\newblock ``Skeleton-based action recognition with shift graph convolutional
  network,''
\newblock in {\em Proceedings of the IEEE/CVF Conference on Computer Vision and
  Pattern Recognition}, 2020, pp. 183--192.

\bibitem{ye2020dynamic}
Fanfan Ye, Shiliang Pu, Qiaoyong Zhong, Chao Li, Di~Xie, and Huiming Tang,
\newblock ``Dynamic gcn: Context-enriched topology learning for skeleton-based
  action recognition,''
\newblock in {\em Proceedings of the 28th ACM International Conference on
  Multimedia}, 2020, pp. 55--63.

\bibitem{song2020stronger}
Yi-Fan Song, Zhang Zhang, Caifeng Shan, and Liang Wang,
\newblock ``Stronger, faster and more explainable: A graph convolutional
  baseline for skeleton-based action recognition,''
\newblock in {\em Proceedings of the 28th ACM International Conference on
  Multimedia}, 2020, pp. 1625--1633.

\end{thebibliography}

\end{document}